\documentclass[conference]{IEEEtran}
\usepackage[pdftex]{graphicx}
\graphicspath{{images/}}

\usepackage{cite}
\usepackage{amsmath}
\usepackage{array}
\usepackage{mdwmath}
\usepackage{url}
\usepackage{stfloats}

\makeatletter
\let\MYcaption\@makecaption
\makeatother

\usepackage[font=footnotesize]{subcaption}

\makeatletter
\let\@makecaption\MYcaption
\makeatother

\begin{document}

\IEEEoverridecommandlockouts


\title{Image Dehazing via Joint Estimation of Transmittance Map and Environmental Illumination}
\author{\IEEEauthorblockN{Sanchayan Santra, Ranjan Mondal}
\IEEEauthorblockA{Indian Statistical Institute, Kolkata, India\\
Email: sanchayan\_r@isical.ac.in, ranjan.rev@gmail.com}
\and
\IEEEauthorblockN{Pranoy Panda, Nishant Mohanty, Shubham Bhuyan}
\IEEEauthorblockA{National Institute of Technology, Rourkela, India\\
Email: \{115ei0353, 115ei0349\}@nitrkl.ac.in,\\ shubhambhuyan07@gmail.com}
}

\maketitle

\begin{abstract}
Haze limits the visibility of outdoor images, due to the existence of fog, smoke and dust in the atmosphere. Image dehazing methods try to recover haze-free image by removing the effect of haze from a given input image. In this paper, we present an end to end system, which takes a hazy image as its input and returns a dehazed image. The proposed method learns the mapping between a hazy image and its corresponding transmittance map and the environmental illumination, by using a multi-scale Convolutional Neural Network. Although most of the time haze appears grayish in color, its color may vary depending on the color of the environmental illumination. Very few of the existing image dehazing methods have laid stress on its accurate estimation. But the color of the dehazed image and the estimated transmittance depends on the environmental illumination. Our proposed method exploits the relationship between the transmittance values and the environmental illumination as per the haze imaging model and estimates both of them. Qualitative and quantitative evaluations show, the estimates are accurate enough. 
 
\end{abstract}

\section{Introduction}
\begin{figure}
\centering
\includegraphics[width=0.48\linewidth]{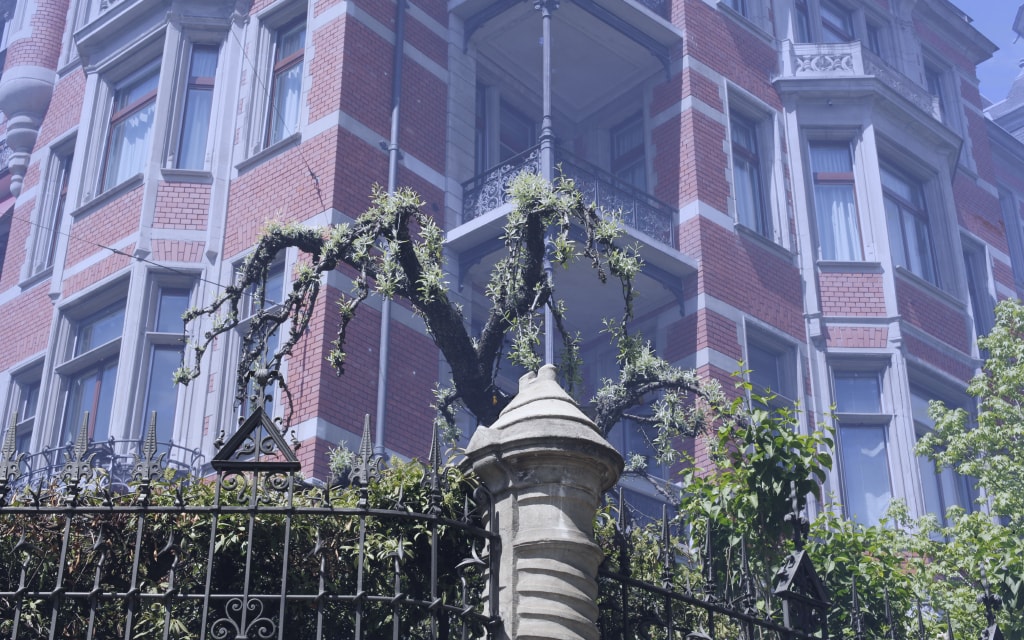}
\includegraphics[width=0.48\linewidth]{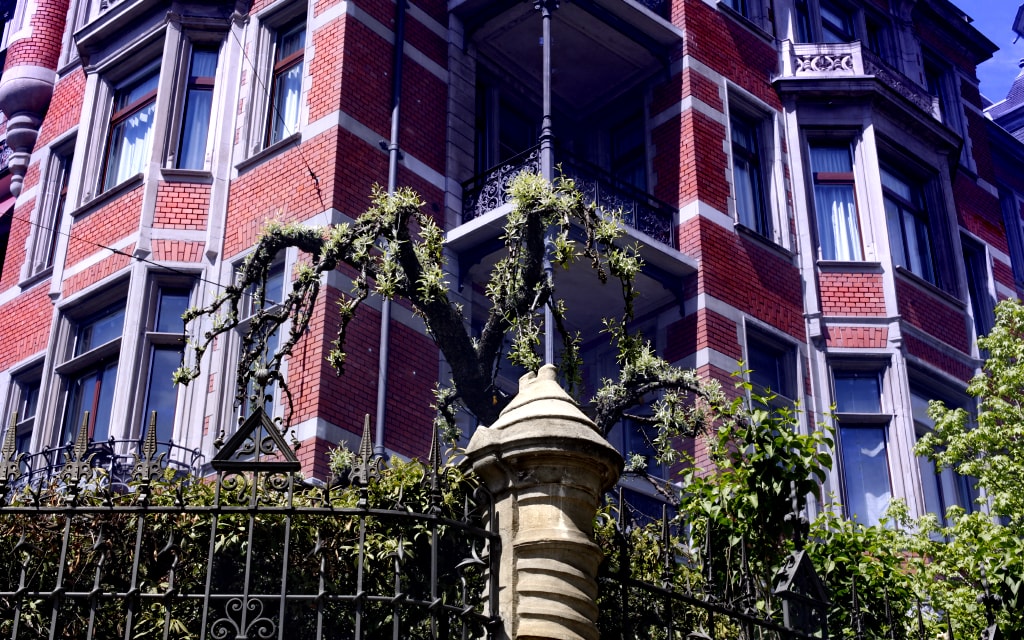}
\caption{Hazy image and its dehazed version obtained by our method}
\label{fig:dehaze_example}
\end{figure}

Haze is an atmospheric phenomenon where the entire scene is obscured by mist, fog, dust or smoke. Haze often occurs when dust and smoke particles accumulate in relatively dry air and scatter light in the environment. This scattering of light by the particles present in the atmosphere, reduces the scene radiance reaching the camera or the observer and adds another layer of the surrounding scattered light, known as the airlight. The attenuated radiance causes the intensity from the scene to get weaker, while the airlight causes the scene to look translucent: sometimes whitish, sometimes colored. Existence of haze in the atmosphere reduces visibility and obstructs the view of distant objects. The problem of the visibility reduction due to haze is common nowadays due to increased use of fossil fuels and smoke from the industries. Image dehazing methods try to recover a haze free version of a given hazy image (Fig \ref{fig:dehaze_example}). In the modern era where smoke and dust are a significant part of our atmosphere, the problem of image dehazing is even more pertinent to be solved. Moreover, methods to solve many real world computer vision and image processing problems, such as surveillance and tracking, suffer due to degradation in visibility. Since most of them assume clear scenes under good weather condition, addressing this problem via image dehazing method is therefore of practical importance.
In this paper, we propose a method for recovering the haze-free image, given a single hazy image as input. We use a single multi-scale Convolutional Neural Network (CNN) for estimating both the transmittance map and the environmental illumination. We have used NYU Depth dataset \cite{silberman2012nyu} to synthetically generate the training data for our network. Although this dataset is an indoor image dataset, the results show this is not an hindrance and our method achieves good results in both synthetic and real world images.

The rest of the paper are organized as follows: Section \ref{sec:prev_work} mentions the prior works on image dehazing with their brief descriptions. How image is formed in haze is described in Section \ref{haze model}, while Section \ref{sec:motivation} describes the idea behind the proposed method. The details of our method is given in Section \ref{sec:method}. In Section \ref{sec:expt_res} we report the experimental setup along with the results obtained by our method. Lastly, Section \ref{sec:conclusion} concludes the paper. 

\IEEEpubidadjcol

\section{Previous Works}
\label{sec:prev_work}
Image Dehazing is a challenging problem to solve. So, the earlier approaches often required multiple images for dehazing the image. These methods \cite{nayar1999vision, narasimhan2002vision} estimated the transmittance map by directly inverting the haze image model. In the recent years many single-image visibility enhancement methods have been proposed \cite{tan2008visibility,he2011single,berman2016non,fattal2014dehazing,ren2016single}. Tan \cite{tan2008visibility} proposed to enhance the visibility of hazy images by maximizing the local contrast, but the enhanced images often contained distorted colors. He \emph{et al.}  \cite{he2011single} proposed the dark channel prior which is derived from the characteristic of natural outdoor images. It states that the intensity value of at least one color channel within a local window is close to zero. They state that this is because generally the outdoor images are colorful, i.e. the brightness varies significantly in different color channels. They estimate the transmittance map by using the dark channel prior. Because of its simplicity, many dehazing methods based on the dark channel prior have been proposed \cite{lee2016review}. Although the dark channel prior gives good result in variety of images, it does not perform very well when airlight and the color of objects are similar. Fattal \cite{fattal2014dehazing} proposed a method, with the assumption that the depth value remains nearly constant in a small patch, but the shading can vary. Since the haze model equation is linear in a small patch, the pixels of a patch forms a straight line in the RGB space. This line intersects with the line formed by the airlight. From this point of intersection the transmittance is computed. This method however assumes that the value of environmental illumination is known. Berman \emph{et al.} \cite{berman2016non} have proposed a method based on non-local prior. This method relies on the assumption that colors of a haze-free image can be well approximated by a few hundred distinct colors. They form clusters in the RGB space and pixels in a cluster are often spread in across the image. The presence of haze elongates the shape of each cluster to a line, as each pixel is affected by a different transmittance coefficients due to their unequal distances from the camera. The radius of each cluster center is used in estimating the transmittance. Ren \emph{et al.} \cite{ren2016single} have proposed a method using CNN with a special network design, which estimates the transmittance map via a coarse and fine network. According to the survey by Li \emph{et al.} \cite{li2016haze}, Fattal \cite{fattal2014dehazing} achieves less errors in both transmittance estimation and final dehazing at different haze levels, whereas Tan \cite{tan2008visibility} and He \cite{he2011single} estimate transmittance more accurately as haze level increases. Berman $et$  $al$ \cite{berman2016non} achieves the least error in transmittance at medium haze level, but the error increases as the haze level increases or decreases.

Most of the methods proposed till now have devoted their attention to estimation of scene transmittance. These methods estimate the airlight component of the input image and then compute the transmittance from it. Although this step requires the environmental illumination, not much attention has been given to it barring a few methods \cite{pedone2011robust, sulami2014atmospheric}. Due to this interdependence of these two parameters, in this paper we propose to estimate them jointly. 


\section{Imaging Model for Atmospheric Scattering}
\label{haze model}
Light passing through a scattering medium is attenuated along its original path and is deflected in different directions. This process of scattering of light by particles present in the atmosphere can be modeled by the following equation \cite{koschmieder1925theorie}
\begin{equation}
    \label{eq:1}
    I(\mathbf{x})=J(\mathbf{x})t(\mathbf{x})+(1-t(\mathbf{x}))A,
\end{equation}
\begin{equation}
    \label{eq:2}
    t(\mathbf{x}) = e^{- \beta  d(\mathbf{x})}.
\end{equation}
Where $I(\mathbf{x})$ and $J(\mathbf{x})$ are the observed hazy image and the radiance of the clear scene, $A$ is the global environmental illumination, and $t(\mathbf{x})$ is the scene transmittance describes the proportion of light that reaches the camera. If we assume that the haze is homogeneous, $t(\mathbf{x})$ can be expressed by (\ref{eq:2}), where $\beta$ is the scattering coefficient and $d(\mathbf{x})$ is the scene depth. The second part of the (\ref{eq:1}) models the scattered environmental illumination that also reaches the observer apart from the reflected light. This is termed as airlight and is the reason behind the washed-out appearance of the hazy scenes. The Eq. \eqref{eq:1} is originally defined for measured irradience. But, for image dehazing purpose this is treated as RGB vector equation. $I(\mathbf{x})$, $J(\mathbf{x})$ and $A$ are treated as $3 \times 1$ vectors and $t(\mathbf{x})$ as scalar. Scattering coefficient ($\beta$), in turn $t(\mathbf{x})$, in general depends on the wavelength of the incident light. But as we are dealing with haze and fog, the particle sizes tend to be large compared to the wavelength of light \cite{narasimhan2002vision}. Therefore, we can safely assume $t(\mathbf{x})$ to be independent of wavelength, as a result same for all the channels. 




\begin{figure*}
\centering
\includegraphics[width=0.9\linewidth]{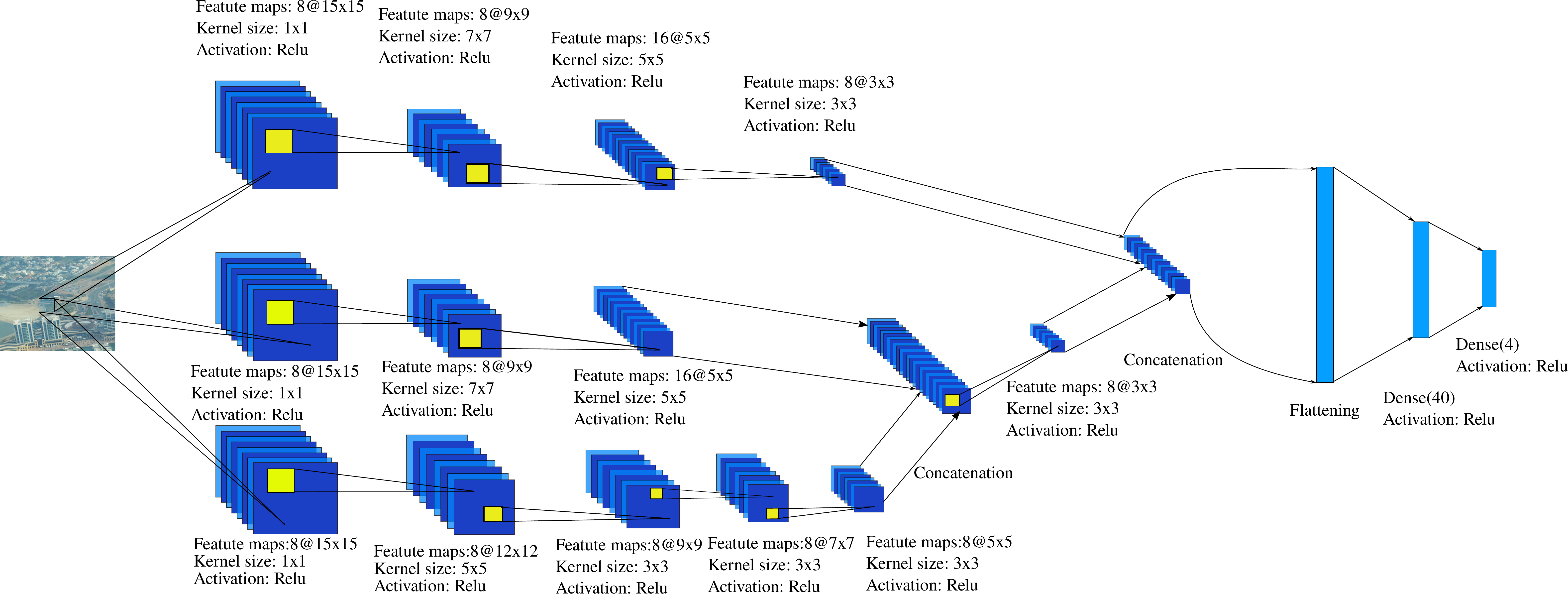}
\caption{The architecture of our joint estimator}
\label{fig:cnn_model}
\end{figure*}
\section{Proposed Approach}
\label{sec:motivation}
Recovering the clear scene radiance from the input hazy image is an ill posed problem, as we have three unknowns out of the four variables present in the haze formation model (Eq. \eqref{eq:1}). So, we have made some simplifying assumptions before solving the problem. We assume $t(\mathbf{x})$ to be constant in a patch, as depth variation will be negligible if the patch is sufficiently small. Then from each image patch we try to estimate both $t$ and $A$ using the following
\begin{equation}
    I(\mathbf{x})=J(\mathbf{x})t+(1-t)A.
    \label{eq:haze_patch}
\end{equation}
Now, to be able to estimate the parameters from patches, we learn the mapping from $I(\mathbf{x})$'s to $t$ and $A$ using a CNN. So, given a patch this CNN will estimate both $t$ and $A$. Now to dehaze a given image we first break it into overlapping patches and then from each patch we estimate $t$ and $A$. We aggregate the $t$ estimates to form transmittance map and $A$'s to a single environmental illumination. Then we use Eq. \eqref{eq:1} to obtain the dehazed image. 


\subsection{Joint $t$ and `$A$' Estimator}
\label{sec:cnn_est}
In our model, as shown in Fig \ref{fig:cnn_model}, we are computing features in three different paths. First path consists of 8 kernels of size 1 $\times$ 1, then 8 kernels of size 5 $\times$ 5. This is followed by 4  layers, each of them has 8 kernels of size 3 $\times$ 3. The path in the middle consists 8 kernels of size 1 $\times$ 1, then followed by 8 kernels of size 7 $\times$ 7 and then 16 kernels of size 5 $\times$ 5. Then, output feature maps of the bottom network and the middle network are concatenated and convolved with 8 kernels of size 3 $\times$ 3. The upper most layer consists 8 kernels of size 1 $\times$ 1, followed by 8 kernels of size 7 $\times$ 7 and then 16 kernels of size 5 $\times$ 5 and then 8 kernels of size 3 $\times$ 3. The output of the upper most layer and the output of other network is then concatenated and flattened. After the flattening we have a dense layer with 40 neurons in it. Then finally the output is obtained from the 4 neurons (one for $t$ and 3 for `$A$') following the dense layer. We have used ReLU (Rectified Linear Unit) as the nonlinear activation function after each layer. The idea behind the model was to take fine details with small convolution kernels and coarse details using the bigger convolution kernels. The model has two layers with bigger convolution kernels and one layer with small convolution kernels. They were added knowing the requirement of a single model to determine both the parameters ($t$ and $A$) as the parameters are dependent on each other through a single equation (Eq. \ref{eq:1}). 



\section{Dehazing Method}
\label{sec:method}
The proposed method takes the following steps in order to dehaze a given image.
\begin{enumerate}
   \item Estimation of transmittance and environmental illumination from patches
   \item Aggregation and creation of transmittance map
   \item Recovering the scene radiance
\end{enumerate}
First, the input image is divided into overlapping patches. From each patch we estimate its corresponding $t$ and $A$ using our estimator network (Sec. \ref{sec:cnn_est}). Then from each estimate we create the transmittance map for the whole image and the global environmental illumination. Then we invert the haze model to recover the scene radiance. These steps are discussed in details in the following subsections. 


\subsection{Estimation of Transmittance and Environmental Illumination from Patches}
We first divide the input image into $15 \times 15$ overlapping patches with stride of 5. Among all these patches, we process only those with intensity variance more than a threshold. Smooth patches are not considered for processing as they do not contain much information. Each one of the selected patch is fed to our joint estimator network and we obtain $t$ and $A$ for each patch. We consider the obtained $t$ as one estimate of transmittance for each pixel of the patch. 

\subsection{Aggregation and Creation of Transmittance Map}
In the previous step of estimating $A$ and $t$, we have taken patches with overlapping pixels. So a pixel will receive multiple estimate of scene transmittanceand we will have many estimates of `$A$'. So, for transmittance, we aggregate them to a single value by taking their average at each pixel. For global atmospheric light we take average all the $A$'s we obtain from different patches. After aggregation it is quite likely that at some pixels estimate of $t$ is not computed. This can happen as we have discarded some patches while estimating $t(\mathbf{x})$. But we require transmittance at each pixel to dehaze an image. So, we interpolate values at those places to create the transmittance map. This is done by minimizing the following function \cite{fattal2014dehazing},
\begin{equation}
\psi(t(\mathbf{x})) = \sum_{\mathbf{x}} s(\mathbf{x})(t(\mathbf{x}) - \tilde{t}(\mathbf{x}))^2 + \lambda \sum_{\mathbf{x}}\sum_{y\in N(\mathbf{x})} \frac{(t(\mathbf{x}) - t(\mathbf{y}))^2}{||I(\mathbf{x}) - I(\mathbf{y})||^2}.
\end{equation}
Where $\tilde{t}(\mathbf{x})$ is the estimated and aggregated scene transmittance. $t(\mathbf{x})$ is the interpolated transmittance that we are trying to obtain. $s(\mathbf{x})$ is either 1 or 0 depending on whether we have an estimate of transmittance at pixel $\mathbf{x}$ or not. $N(\mathbf{x})$ denotes the four neighbors of pixel $\mathbf{x}$. 


\subsection{Recovering the scene radiance}
We have already obtained $t(\mathbf{x})$ at each pixel and `$A$' for the entire image. So, we can recover the scene radiance of the input hazy image using \eqref{eq:1}. But this can amplify noise in the recovered image, especially in the regions with dense haze. Therefore, we restrict the transmittance by a lower bound of 0.1 and recover the image by the following:
\begin{equation}
J(\mathbf{x}) = A + \frac{I(\mathbf{x}) - A}{\max{\{0.1, t(\mathbf{x})\}}}.
\end{equation}

\section{Experimental Results}
\label{sec:expt_res}
\subsection{Data Generation for training the network}
\begin{figure}
    \centering     
    \begin{subfigure}[t]{0.23\linewidth}
        \includegraphics[width=\linewidth]{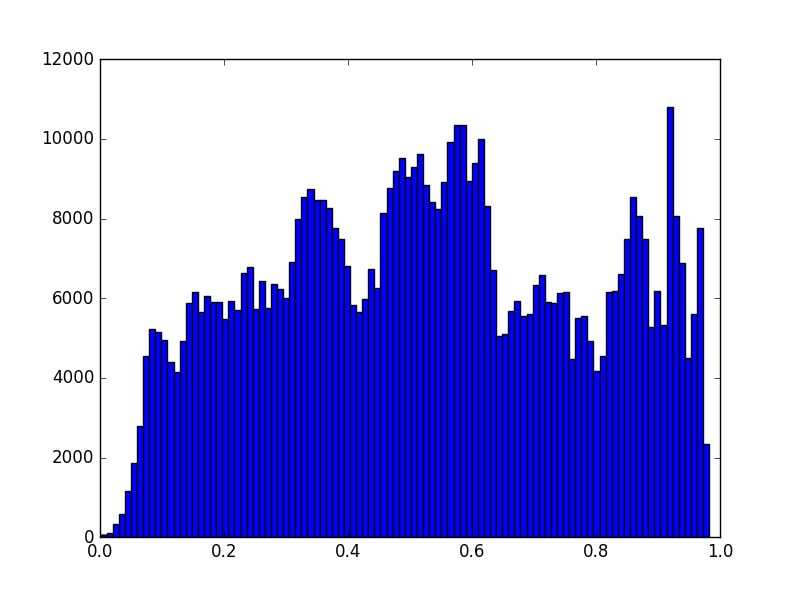}
        \caption{t}
        \label{fig:a}
    \end{subfigure}
    \begin{subfigure}[t]{0.23\linewidth}
        \includegraphics[width=\linewidth]{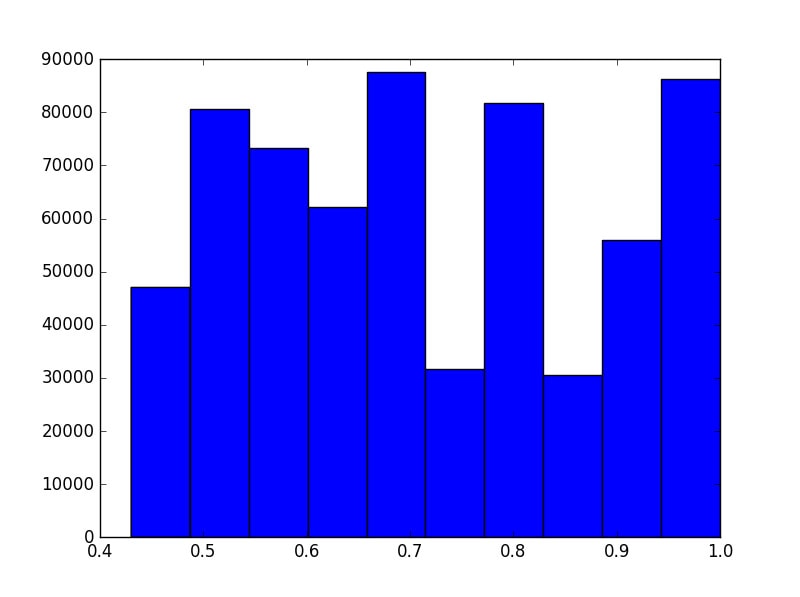}
        \caption{A-Red}
        \label{fig:b}
    \end{subfigure}
    \begin{subfigure}[t]{0.23\linewidth}
        \includegraphics[width=\linewidth]{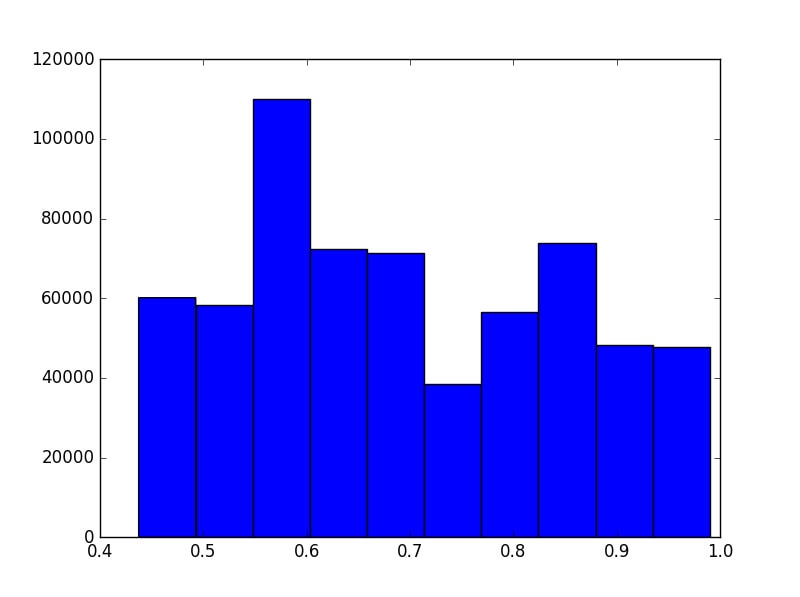}
        \caption{A-Green}
        \label{fig:c}
    \end{subfigure}
    \begin{subfigure}[t]{0.23\linewidth}
        \includegraphics[width=\linewidth]{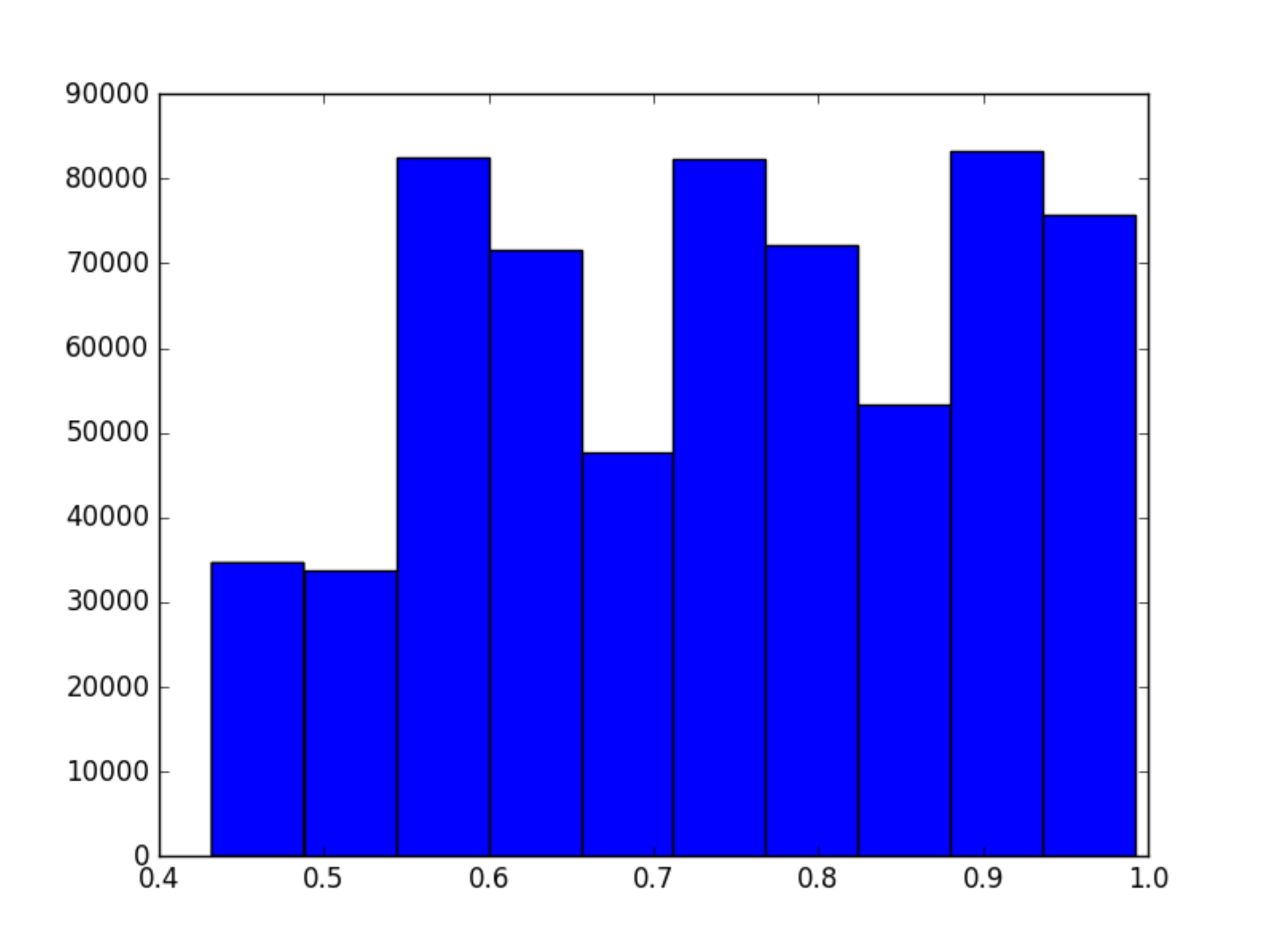}
        \caption{A-Blue}
        \label{fig:d}
    \end{subfigure}
    \caption{Histogram of the generated transmittance values and each of the RGB channel of environmental illumination}
    \label{fig:hist}
\end{figure}
For the purpose of training the network we have generated hazy images using clean images along with their depth maps from NYU depth dataset \cite{silberman2012nyu}. We generate hazy images by applying Eq. \eqref{eq:1} and \eqref{eq:2} with different values of environmental illumination and scattering coefficient ($\beta$). The value of the transmittance $t(\mathbf{x})$lies between 0 and 1,and it depends both on the depth at that point as well as on $\beta$. As our dataset contains only indoor images, the variation of depth in an image is not much. So in order to make the training data for $t(\mathbf{x})$ estimation diverse, we uniformly varied $\beta$ between 0.5 and 1. The distribution is shown in Fig. {\ref{fig:a}}. Similarly we generate the $A$ values for the training data and Fig. \ref{fig:b}, \ref{fig:c}, \ref{fig:d} shows the distribution of the obtained values. Now, from the generated hazy images, we extract RGB patches of size $15 \times 15$ with a stride of 5 pixels. We don't use all of the extracted patches, as not all of them are good enough for training purpose. We discard smooth patches because they have very little information in them. So, if the variance of a patch lies below a threshold, then that patch is discarded. In the depth maps of NYU dataset, depth information is not present for all pixels. So if the depth information of many pixels in a particular patch is not present, then that patch is also discarded. Patches are discarded in order to ensure that the model gets patches with proper information in them.

\subsection{Experimental Settings}
Learning the mapping between hazy images and corresponding transmittance map and A values is accomplished by minimizing the loss between the output and the corresponding ground truth. The output is a vector consisting of the transmittance value and $A$ values of that particular patch. We have used mean squared error as the loss function. We have used Adadelta optimizer to train the network with a batch size of 1000. The network was trained for 90 epochs.





\subsection{Quantitative Evaluation}
\begin{figure}
    \centering
    \begin{subfigure}[t]{0.19\linewidth}
        \centering
        \includegraphics[width=\linewidth]{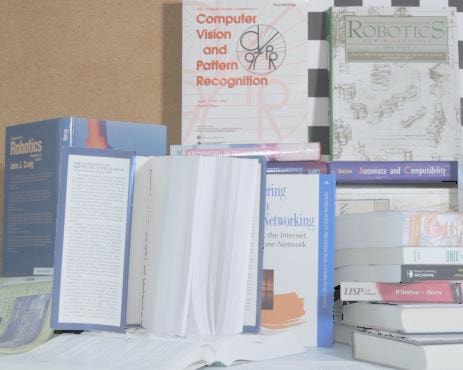}\\\vfill
        \includegraphics[width=\linewidth]{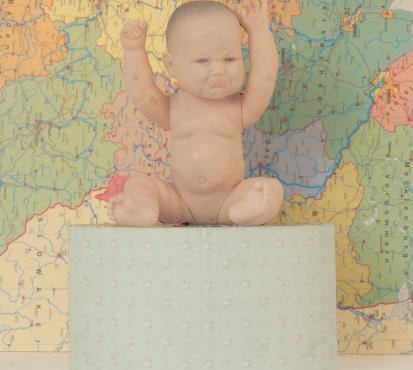}\\\vfill
        \includegraphics[width=\linewidth]{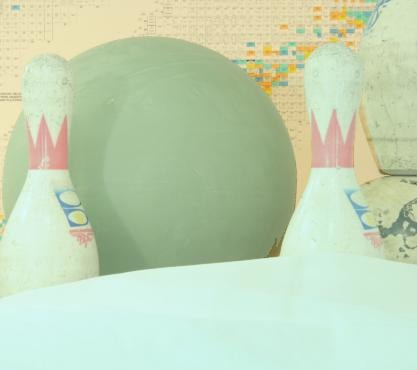}\\\vfill
        \includegraphics[width=\linewidth]{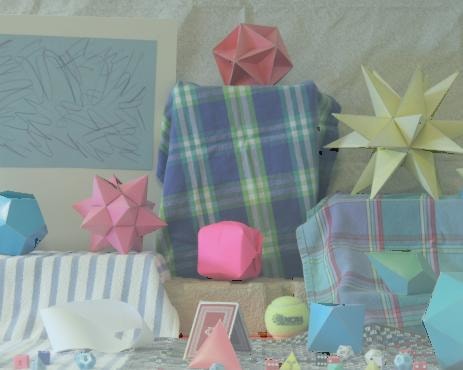}
        \caption{Input}
        \label{fig:mb_inp}
    \end{subfigure}
    \begin{subfigure}[t]{0.19\linewidth}
        \centering
        \includegraphics[width=\linewidth]{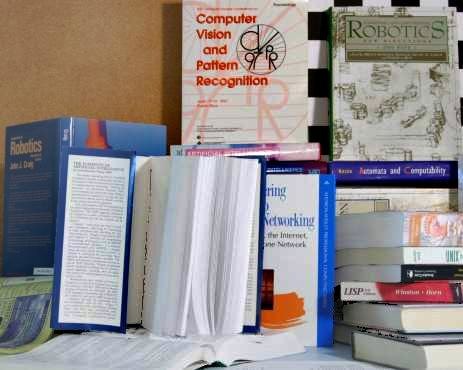}\\\vfill
        \includegraphics[width=\linewidth]{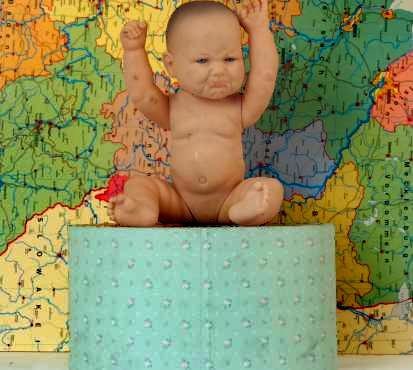}\\\vfill
        \includegraphics[width=\linewidth]{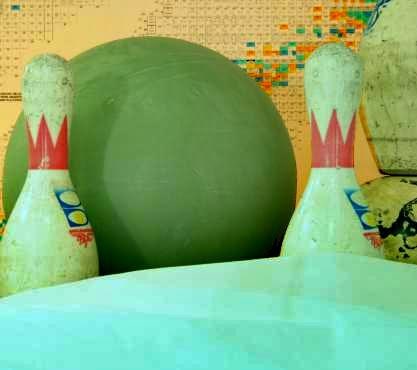}\\\vfill
        \includegraphics[width=\linewidth]{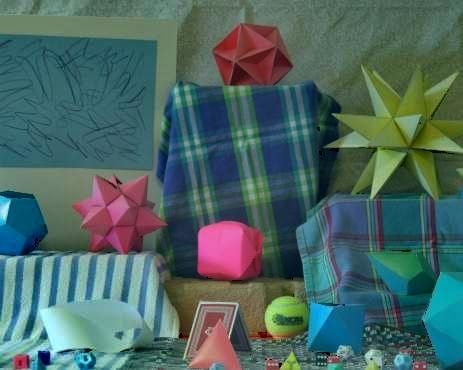}
        \caption{Ren \emph{et al.}}
        \label{fig:mb_ren}
    \end{subfigure}
    \begin{subfigure}[t]{0.19\linewidth}
        \centering
        \includegraphics[width=\linewidth]{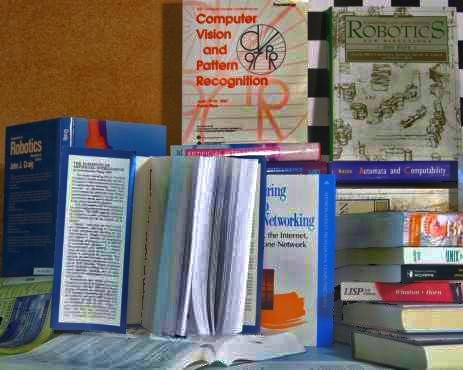}\\\vfill
        \includegraphics[width=\linewidth]{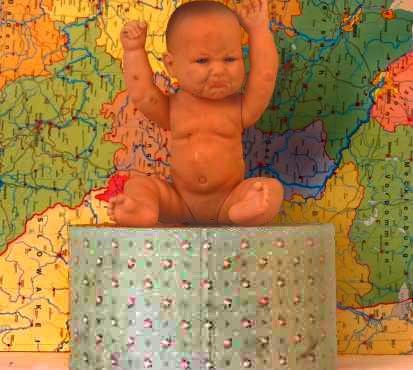}\\\vfill
        \includegraphics[width=\linewidth]{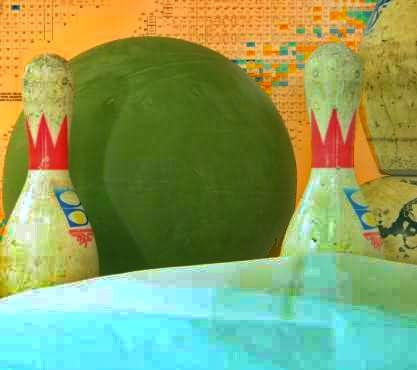}\\\vfill
        \includegraphics[width=\linewidth]{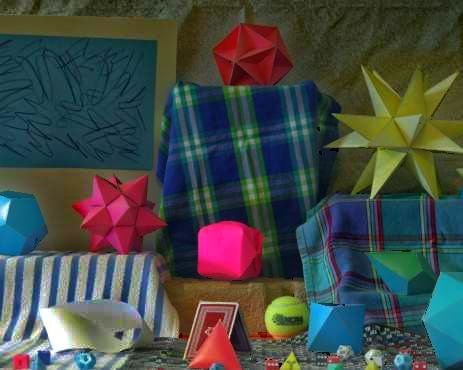}
        \caption{He \emph{et al.}}
        \label{fig:mb_dcp}
    \end{subfigure}
    \begin{subfigure}[t]{0.19\linewidth}
        \centering
        \includegraphics[width=\linewidth]{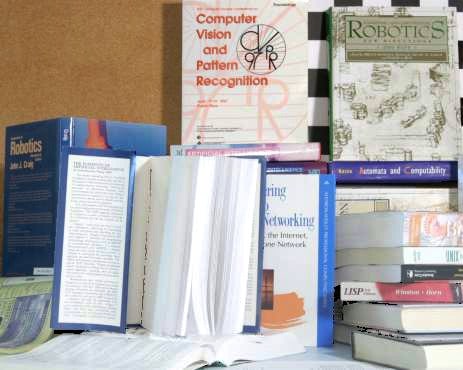}\\\vfill
        \includegraphics[width=\linewidth]{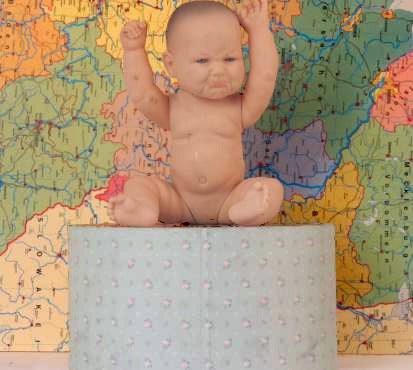}\\\vfill
        \includegraphics[width=\linewidth]{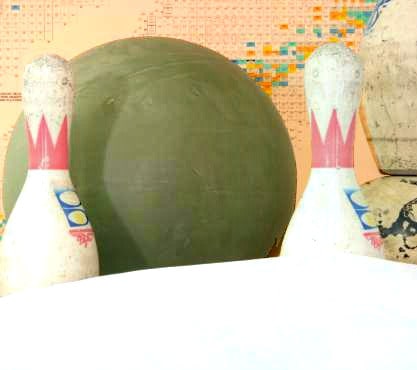}\\\vfill
        \includegraphics[width=\linewidth]{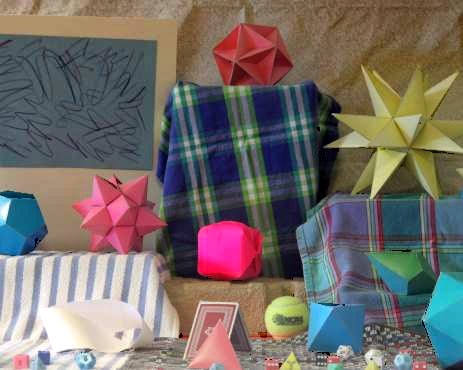}
        \caption{Our}
        \label{fig:mb_our}
    \end{subfigure}
    \begin{subfigure}[t]{0.19\linewidth}
        \centering
        \includegraphics[width=\linewidth]{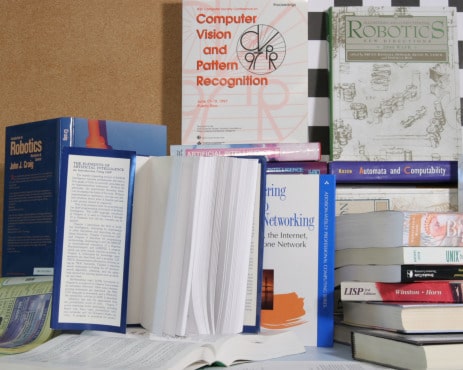}\\\vfill
        \includegraphics[width=\linewidth]{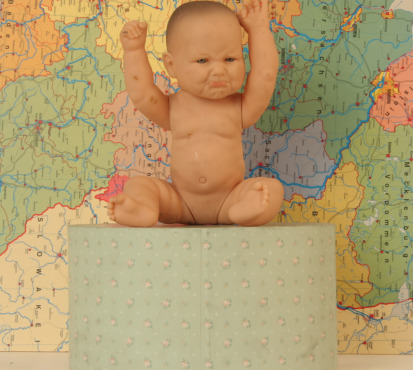}\\\vfill
        \includegraphics[width=\linewidth]{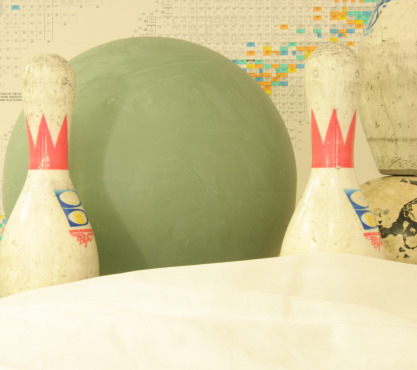}\\\vfill
        \includegraphics[width=\linewidth]{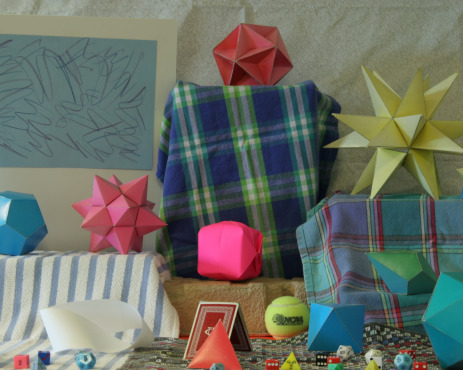}
        \caption{GT}
        \label{fig:mb_gt}
    \end{subfigure}
    \caption{Results on \emph{Books}, \emph{Baby}, \emph{Bowling}, \emph{Moebius}}
    \label{fig:quantitative_eval}
\end{figure}
\begin{table}
    \centering
    \caption{PSNR of dehazed results}
    \label{tab:res_psnr}
    \begin{tabular}{ |c|c|c|c| } 
        \hline
        \textbf{Images} & \textbf{Ren $\textbf{et}$ $\textbf{al.}$ \textbf{\cite{ren2016single}}} & \textbf{He $\textbf{et}$ $\textbf{al.}$ \textbf{\cite{he2011single}}} & \textbf{Ours} \\
        \hline
        $Books$ & \textbf{21.3663} & 14.2198 & 20.3089 \\
        \hline
        $Baby$ & 16.6997 & 16.5203 & \textbf{22.1424} \\
        \hline
        $Bowling$ &12.5822 & 11.0191 & \textbf{18.895} \\
        \hline
        $Moebius$ & 19.5676 & 15.0258 & \textbf{19.9731} \\
        \hline
    \end{tabular}
\end{table}
\begin{table}
    \centering
    \caption{SSIM of dehazed results}
    \label{tab:res_ssim}
    \begin{tabular}{ |c|c|c|c| }  
        \hline
        \textbf{Images} & \textbf{Ren $\textbf{et}$ $\textbf{al.}$ \textbf{\cite{ren2016single}}} & \textbf{He $\textbf{et}$ $\textbf{al.}$ \textbf{\cite{he2011single}}} & \textbf{Ours} \\
        \hline
        $Books$ & 0.9073 & 0.8052 &\textbf{ 0.9280} \\
        \hline
        $Baby$ & 0.8658 & 0.8233 & \textbf{0.9375} \\
        \hline
        $Bowling$ & 0.912 & 0.8612 & \textbf{0.9466} \\
        \hline
        $Moebius$ & \textbf{0.9315} & 0.8339 & 0.9146 \\
        \hline
    \end{tabular}
\end{table}
There is no common agreed metric to quantify the result of a dehazing methods. However, we have used Peak Signal-to-Noise Ratio (PSNR) and Structural Similarity Index (SSIM) to compare the results with the other dehazing methods. We have compared the proposed method with two other dehazing methods that are He \emph{et al.} \cite{he2011single} and Ren \emph{et al.} \cite{ren2016single} using the PSNR (Table \ref{tab:res_psnr}) and SSIM (Table \ref{tab:res_ssim}) metrics.  The hazy image examples were synthesized from the Middlebury Stereo Database \cite{hirschmuller2007evaluation}. We use four examples: \emph{Books}, \emph{Baby}, \emph{Bowlin}, \emph{Moebius} for illustration (Fig. \ref{fig:quantitative_eval}). Fig. \ref{fig:mb_inp} shows the input hazy images generated from clean images with known depth maps. The method of He \emph{et al.} \cite{he2011single} which uses the dark channel prior, tends to overestimate the haze and ends up making the results darker (Fig. \ref{fig:mb_dcp}). The dehazed results by Ren \emph{et al.} \cite{ren2016single} method (Fig. \ref{fig:mb_ren}) satisfactorily removes haze in most of the cases. The dehazed results obtained by the proposed method (Fig. \ref{fig:mb_our}) are quite close to the ground truths (Fig. \ref{fig:mb_gt}) and end up having relatively higher PSNR and SSIM values in most of the cases. The relatively higher PSNR and SSIM values indicate that the proposed method estimates the transmittance map and the environmental illumination better than the competing dehazing methods.

\subsection{Evaluation on Benchmark Dataset}
\begin{figure}
    \centering     
    \begin{subfigure}[t]{0.32\linewidth}
        \includegraphics[width=\linewidth]{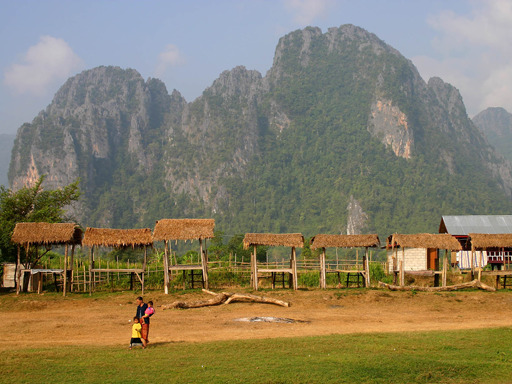}
        \caption{Input}
        \label{fig:mountain_inp}
    \end{subfigure}
    \begin{subfigure}[t]{0.32\linewidth}
        \includegraphics[width=\linewidth]{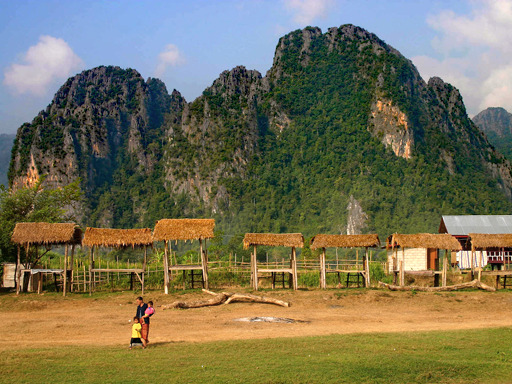}
        \caption{Fattal}
        \label{fig:mountain_cl}
    \end{subfigure}
    \begin{subfigure}[t]{0.32\linewidth}
        \includegraphics[width=\linewidth]{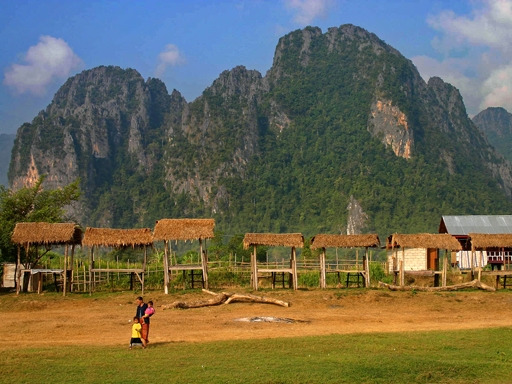}
        \caption{He \emph{et al.}}
        \label{fig:mountain_dcp}
    \end{subfigure}
    
    \begin{subfigure}[t]{0.48\linewidth}
        \includegraphics[width=\linewidth]{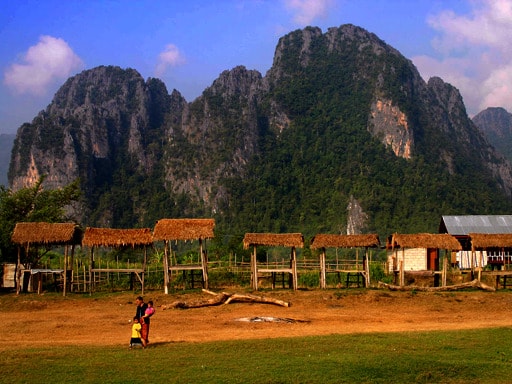}
        \caption{Our}
        \label{fig:mountain_our}
    \end{subfigure}
    \begin{subfigure}[t]{0.48\linewidth}
        \includegraphics[width=\linewidth]{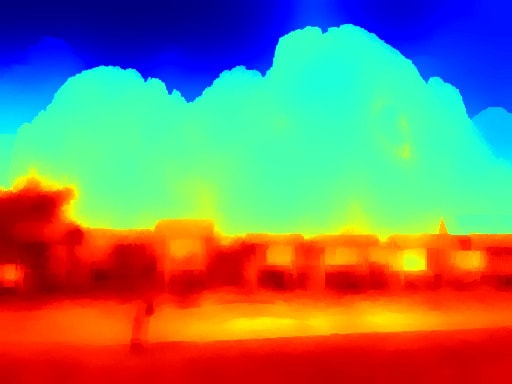}
        \caption{Transmittance Map}
        \label{fig:mountain_t}
    \end{subfigure}
    \caption{\emph{mountain}}
    \label{fig:res_mountain}
\end{figure}
\begin{figure*}
    \centering
    \begin{subfigure}[b]{0.19\linewidth}
        \includegraphics[width=\linewidth]{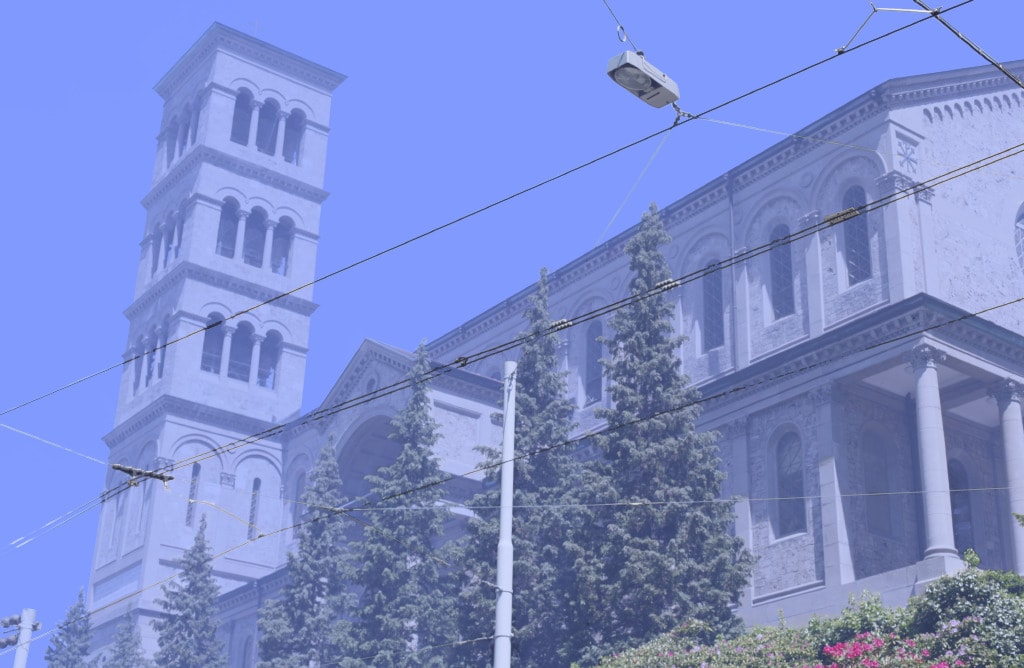}
        \caption{Input}
    \end{subfigure}
    \begin{subfigure}[b]{0.19\linewidth}
        \includegraphics[width=\linewidth]{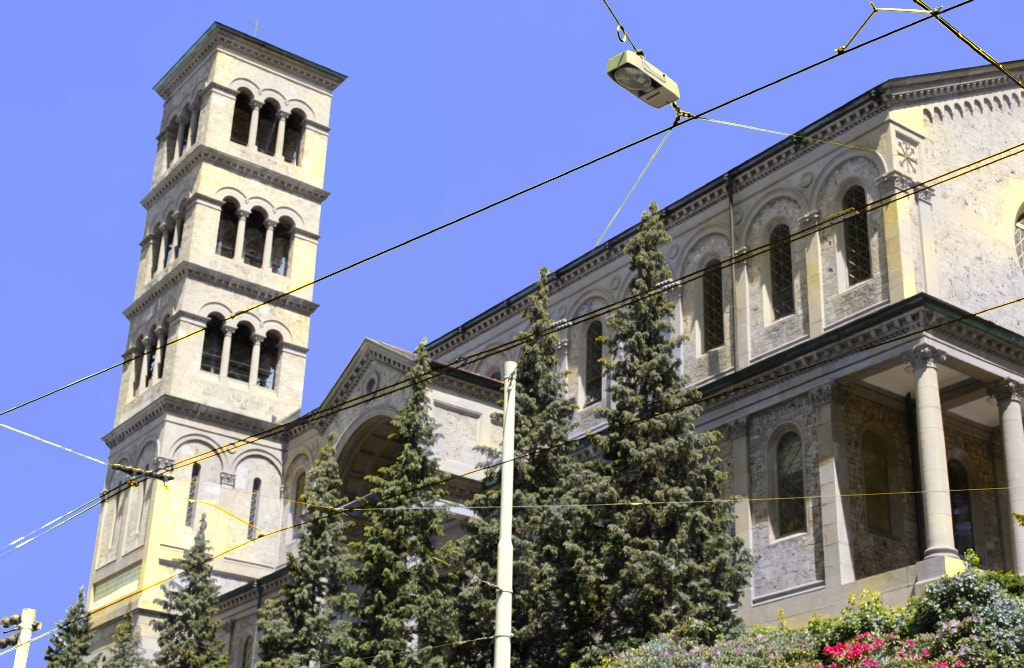}
        \caption{Fattal}
    \end{subfigure}
    \begin{subfigure}[b]{0.19\linewidth}
        \includegraphics[width=\linewidth]{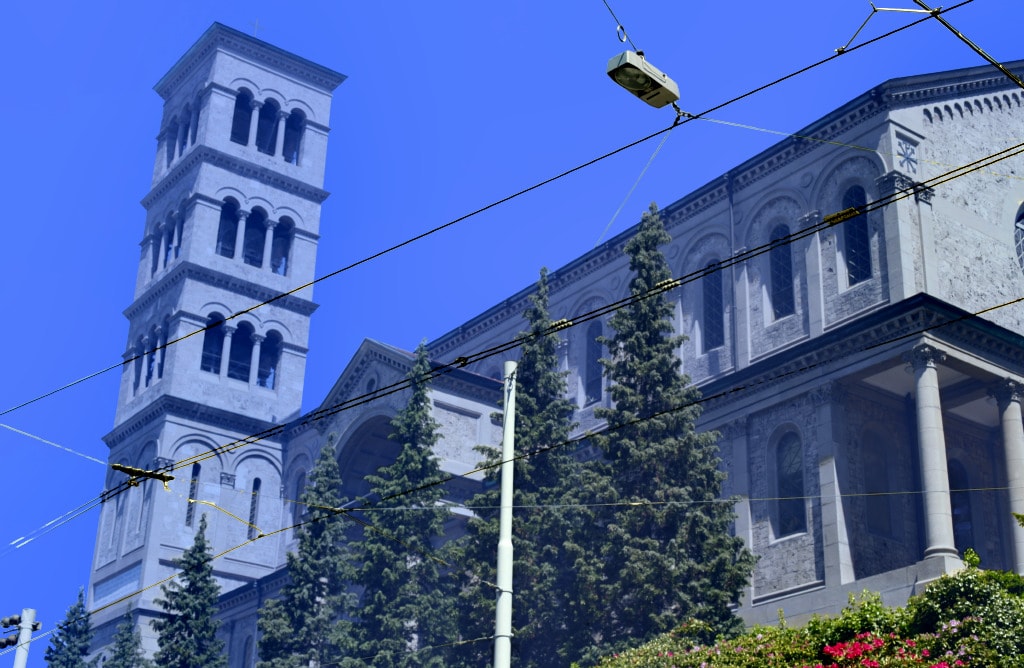}
        \caption{Ren \emph{et al.} \cite{ren2016single}}
    \end{subfigure}
    \begin{subfigure}[b]{0.19\linewidth}
        \includegraphics[width=\linewidth]{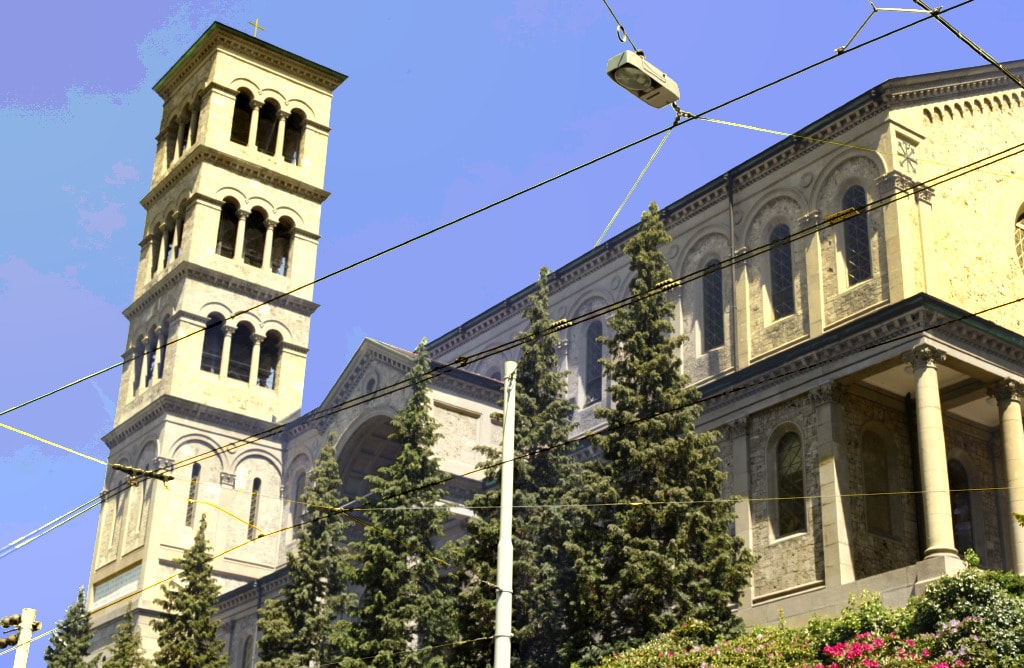}
        \caption{He \emph{et al.}}
    \end{subfigure}
    \begin{subfigure}[b]{0.19\linewidth}
        \includegraphics[width=\linewidth]{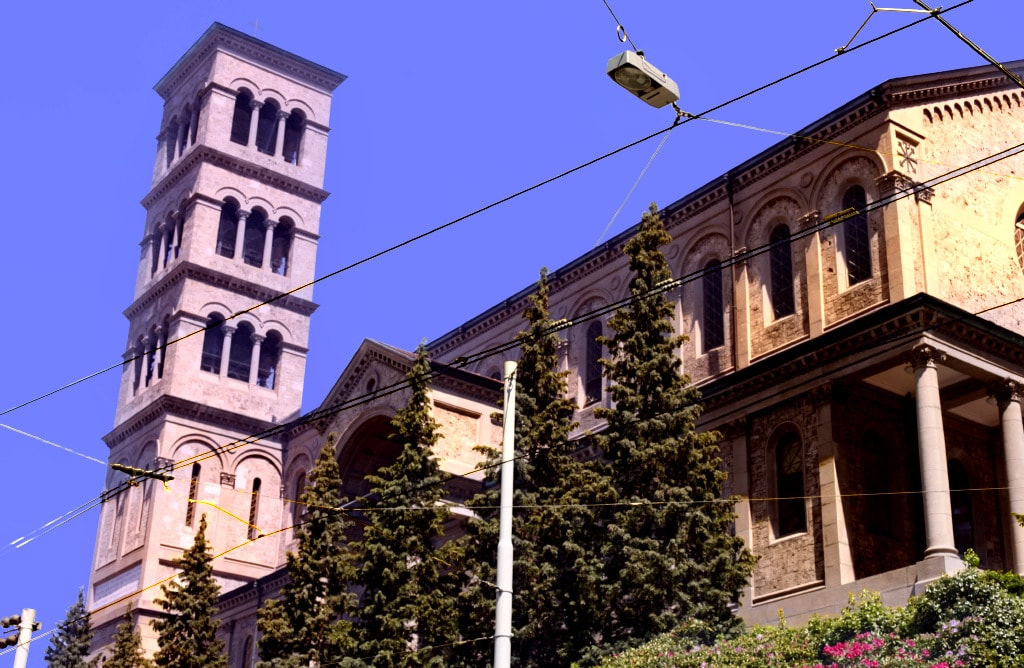}
        \caption{Our}
    \end{subfigure}
    \caption{Church Image}
\end{figure*}
\begin{figure*}
    \centering
    \begin{subfigure}[b]{0.32\linewidth}
        \centering
        \includegraphics[width=\linewidth]{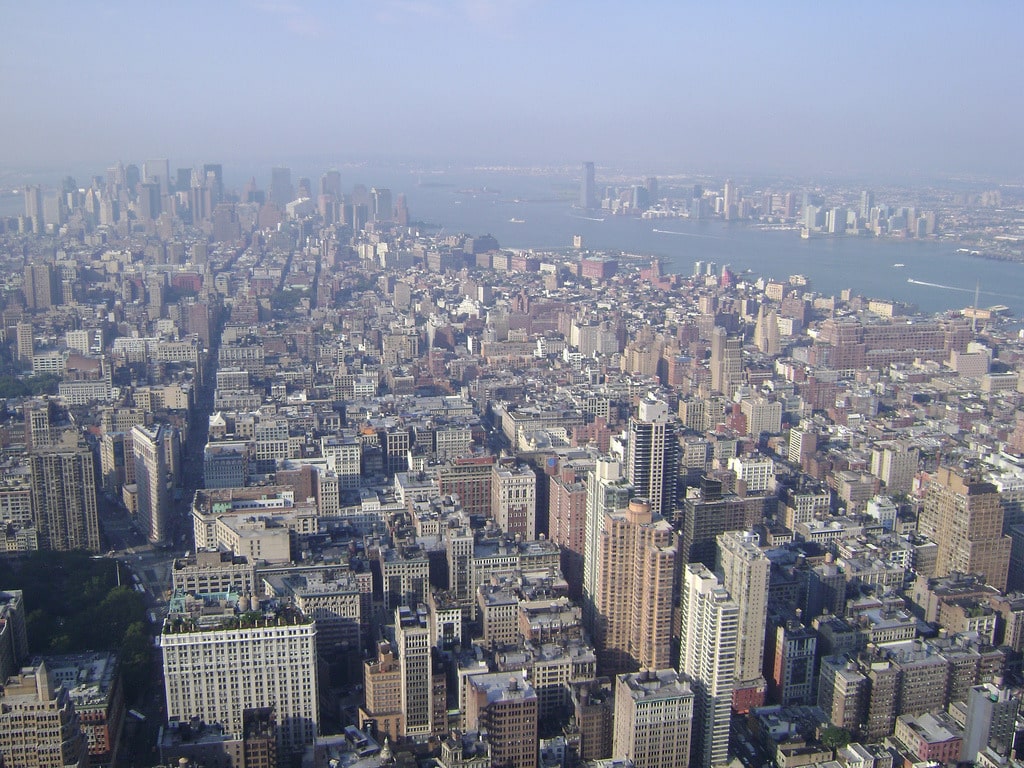}\\\vfill
        \includegraphics[width=\linewidth]{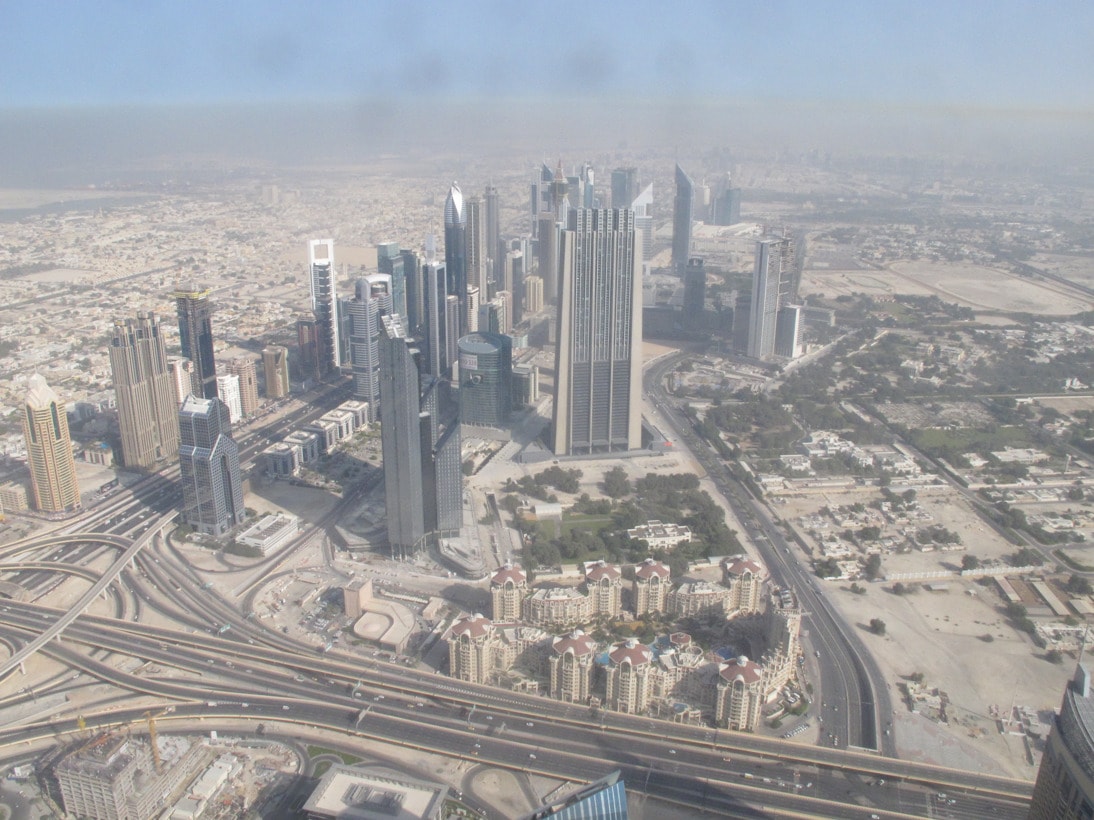}\\\vfill
        \includegraphics[width=\linewidth]{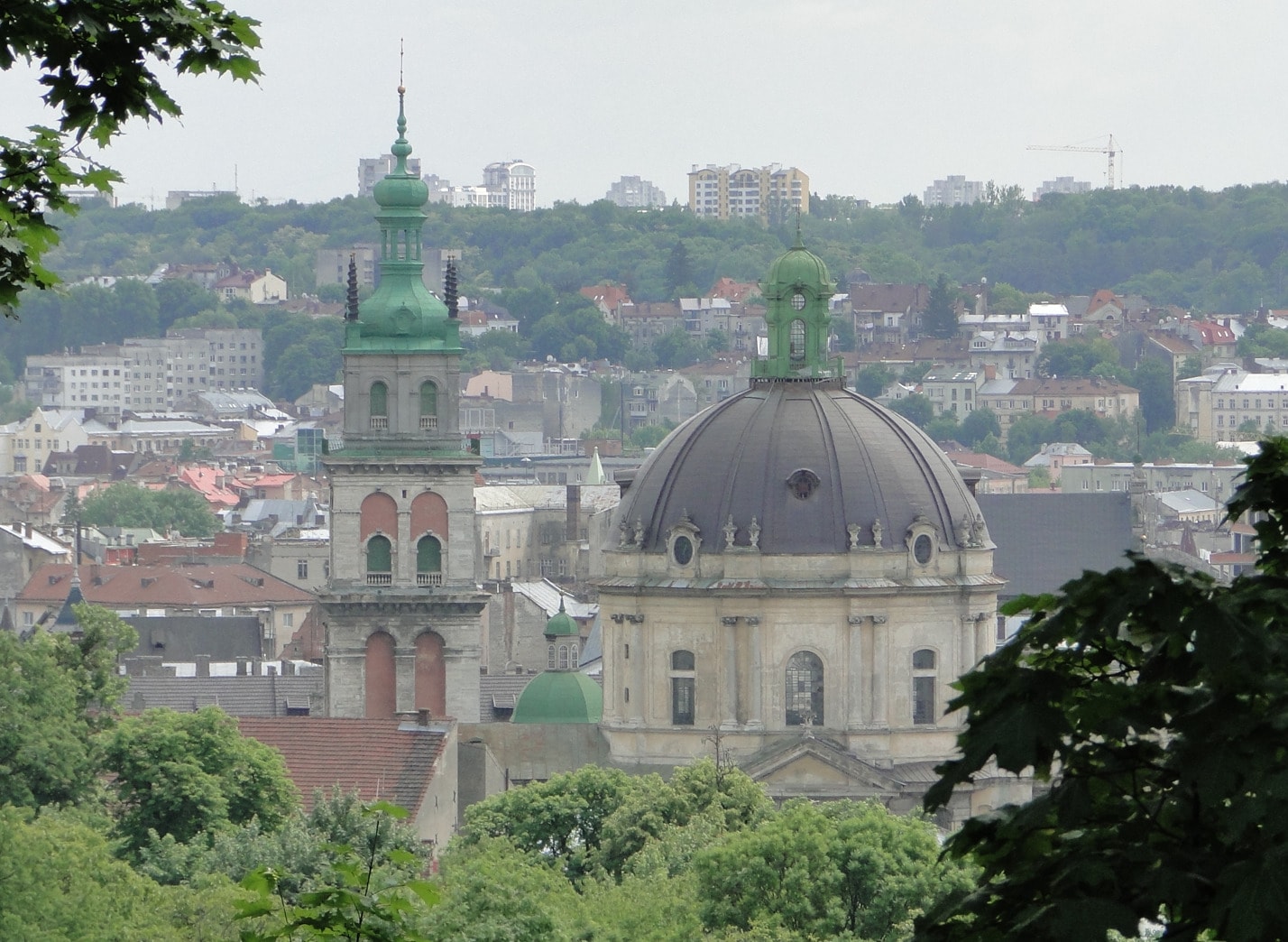}
        \caption{Input}
    \end{subfigure}
    \begin{subfigure}[b]{0.32\linewidth}
        \centering
        \includegraphics[width=\linewidth]{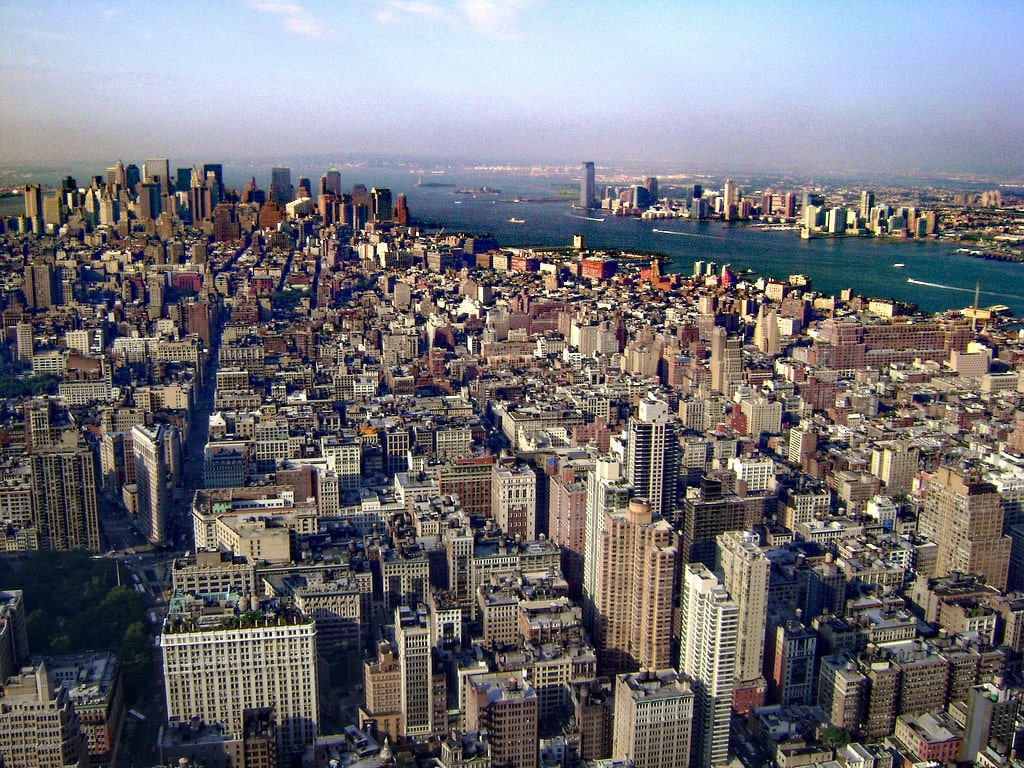}\\\vfill
        \includegraphics[width=\linewidth]{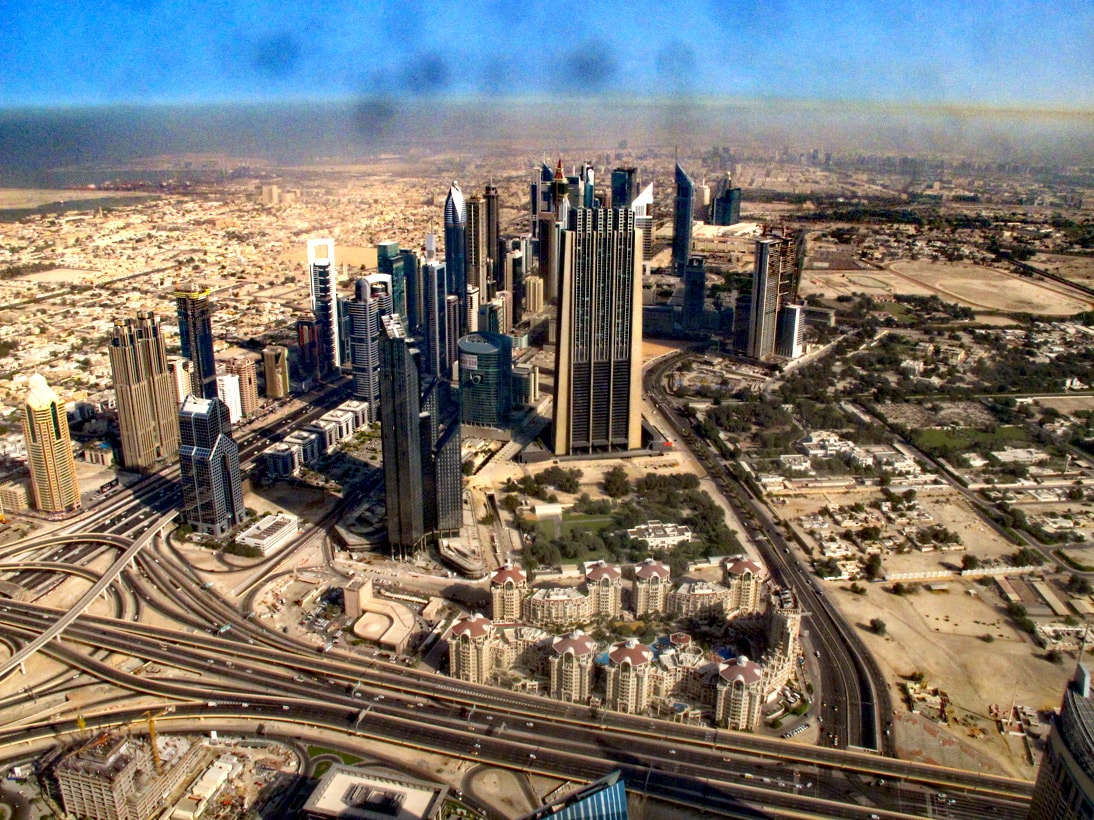}\\\vfill
        \includegraphics[width=\linewidth]{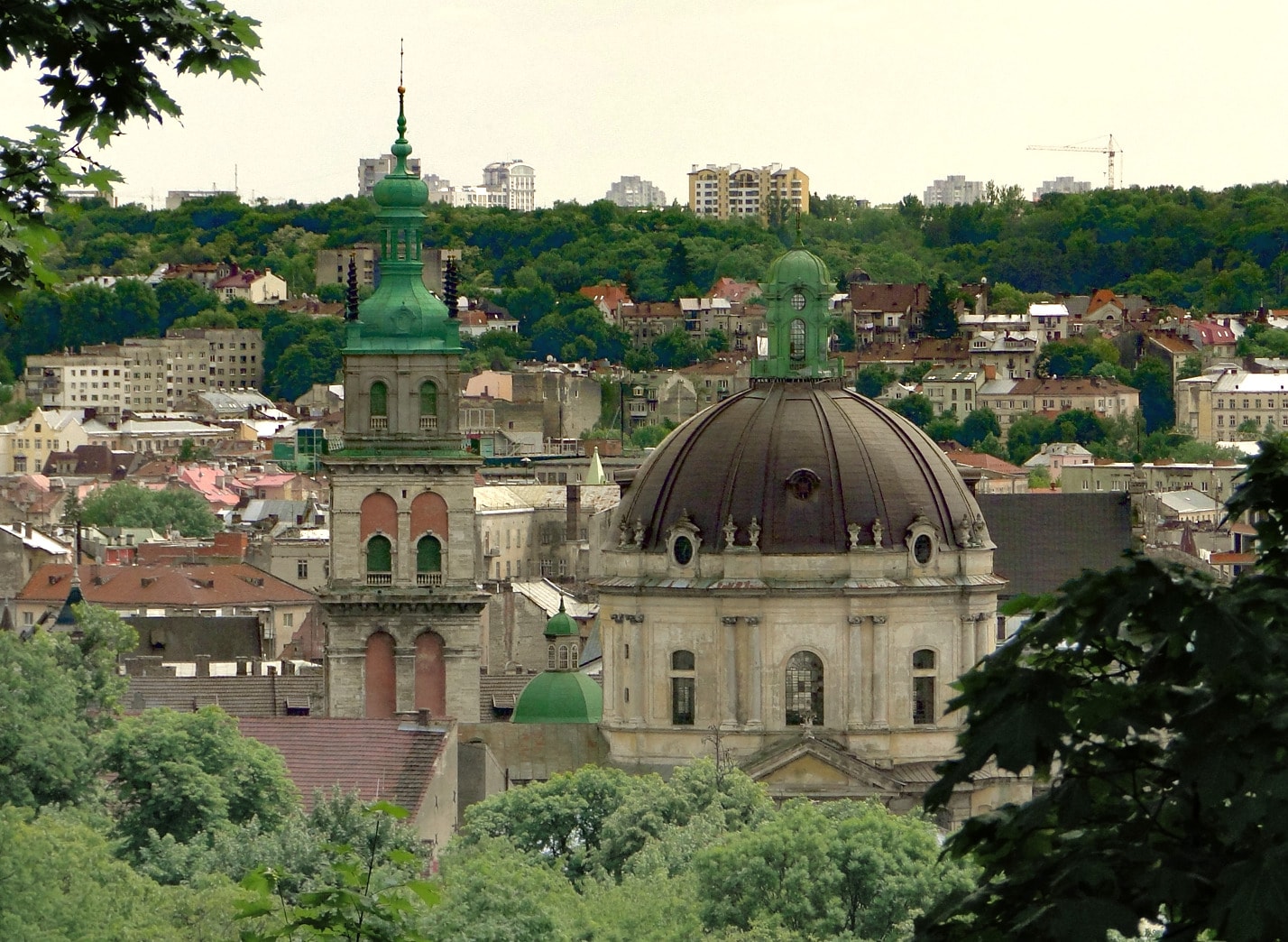}
        \caption{Result of Fattal \cite{fattal2014dehazing}}
    \end{subfigure}
    \begin{subfigure}[b]{0.32\linewidth}
        \centering
        \includegraphics[width=\linewidth]{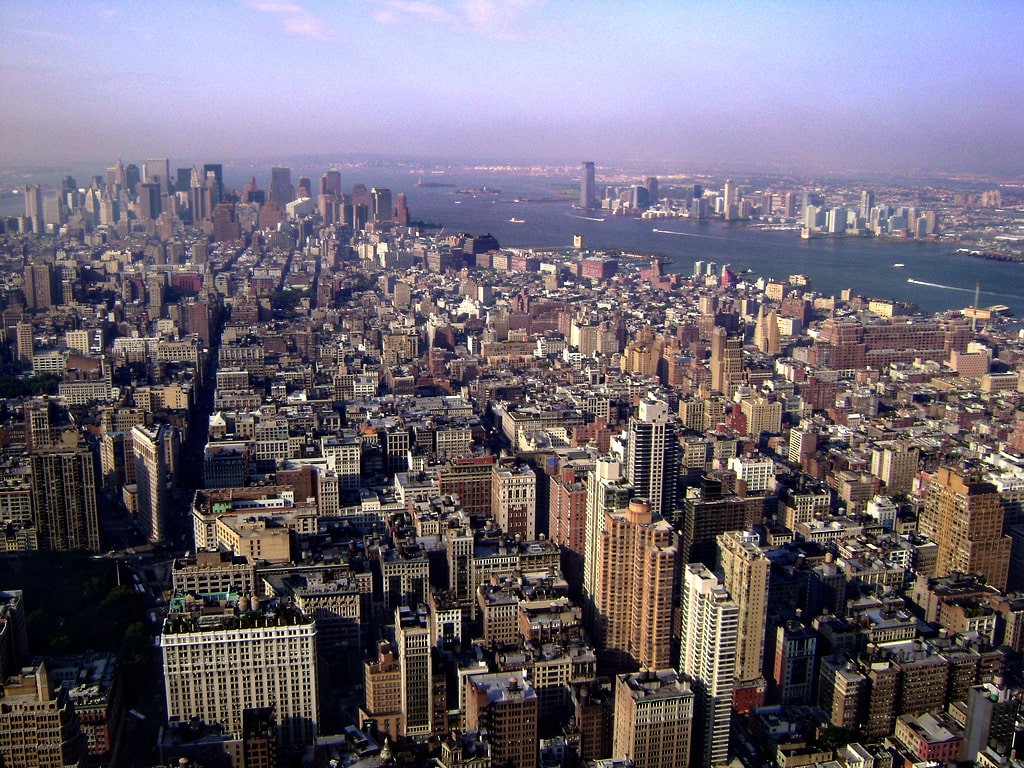}\\\vfill
        \includegraphics[width=\linewidth]{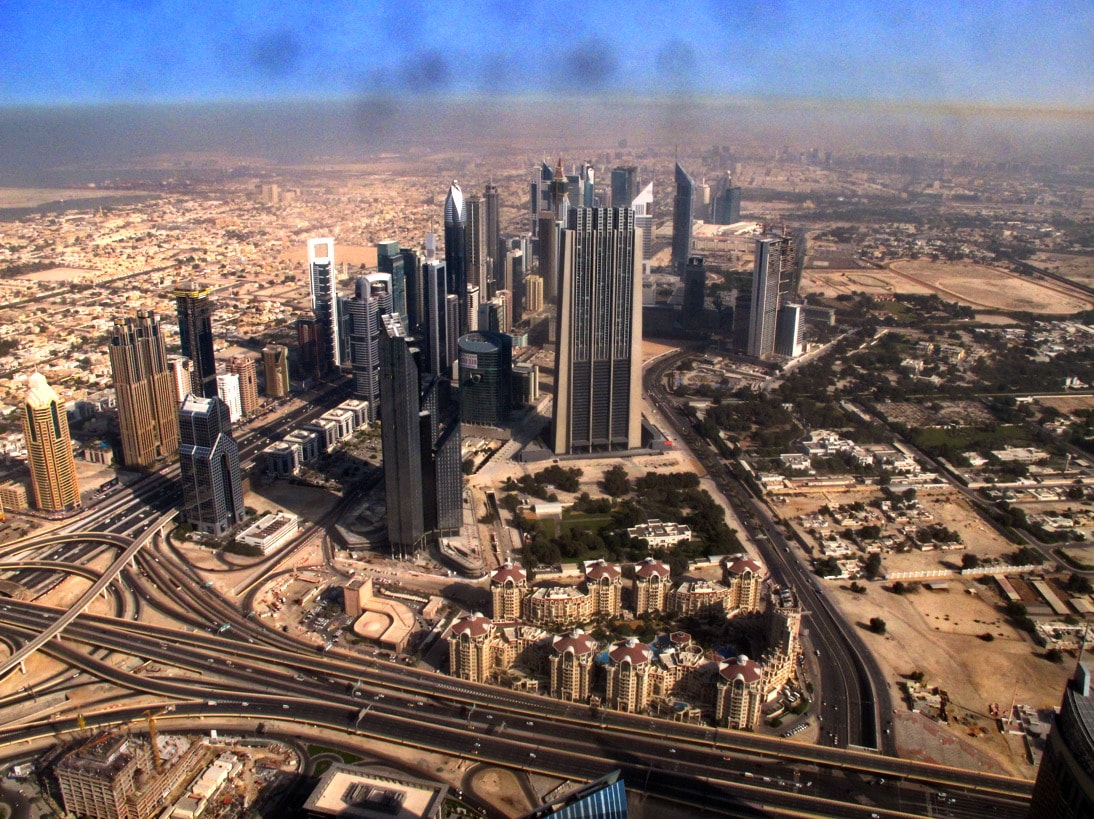}\\\vfill
        \includegraphics[width=\linewidth]{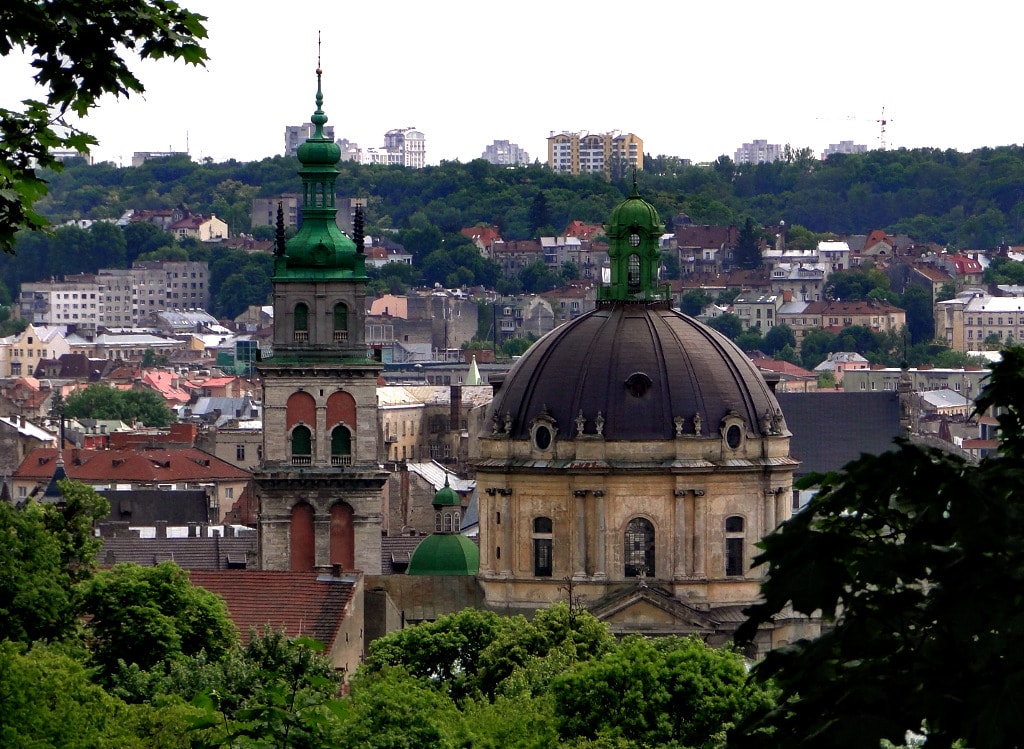}
        \caption{Our result}
    \end{subfigure}
    \caption{Results on \emph{ny17, dubai and lviv}}
    \label{fig:res_outdoor}
\end{figure*}
\begin{figure*}
    \centering
    \begin{subfigure}[t]{0.2\linewidth}
        \includegraphics[width=\linewidth]{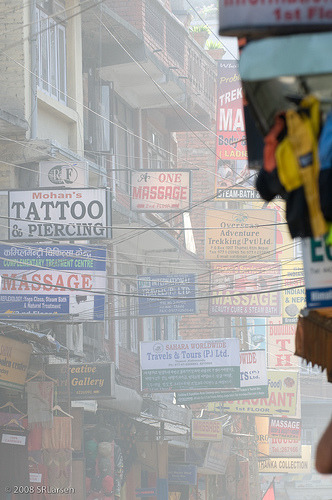}
        \caption{Input}
    \end{subfigure}
        \begin{subfigure}[t]{0.2\linewidth}
        \includegraphics[width=\linewidth]{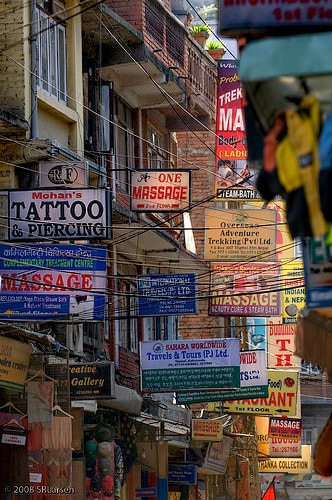}
        \caption{Fattal}
    \end{subfigure}
        \begin{subfigure}[t]{0.2\linewidth}
        \includegraphics[width=\linewidth]{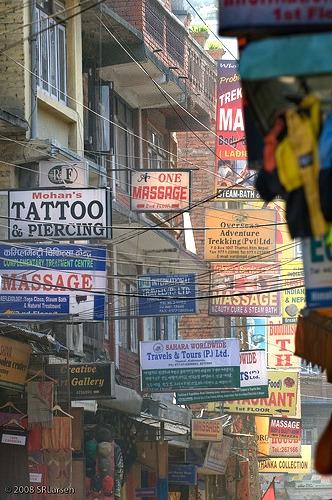}
        \caption{He \emph{et al.}}
    \end{subfigure}
        \begin{subfigure}[t]{0.2\linewidth}
        \includegraphics[width=\linewidth]{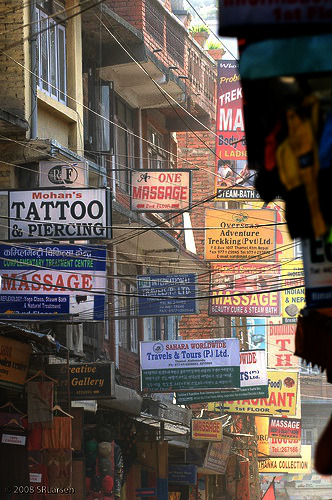}
        \caption{Our}
    \end{subfigure}
    \caption{\emph{flags} image}
    \label{fig:res_flag}
\end{figure*}
We have evaluated our method on a variety of day time images (Fig \ref{fig:res_mountain}, \ref{fig:res_outdoor}, \ref{fig:res_flag}). This includes benchmark images from Fattal's dataset. Although our estimator model is trained hazy images generated from indoor images, it can be applied to dehaze outdoor images. Because, although the depth variation in the indoor images is very less but the variation in transmittance value over the entire training data is diverse. The results on the other benchmark images are shown below for the purpose of qualitative evaluation. 
We have shown the results of our proposed method against Fattal \cite{fattal2014dehazing} and He \emph{et al.} \cite{he2011single}'s methods in Figure \ref{fig:res_mountain}. The dehazed \emph{mountain} image by He \emph{et al.} \cite{he2011single} is a bit on the darker side, due its overestimation of the thickness of the haze. The method of Fattal \cite{fattal2014dehazing} has enhanced the image visibility by removing the haze,  but the value of the environmental illumination used in the dehazed result was manually selected to get the best result. Whereas, the dehazed results obtained by our proposed method (Fig. \ref{fig:mountain_our}) has ended up removing most of the haze, by estimating both the transmittance map (Fig. \ref{fig:mountain_t}) and $A$ value. The correctness of the transmittance map estimated by our proposed method can be inferred from the fact that the transmittance values in the regions near the huts are of high value and the regions near mountains are of low value. This is a correct estimation as the objects near the observer have high transmittance with respect to the objects at a larger distance. The estimated transmittance map has a lot of fine details regarding the shape and structure of the objects in the image. These finer details enables the recovery of a sharp and clear scene radiance. Although, our method performs well on daytime images, it performs poorly on nighttime images. This is due to the fact that the training images were synthesized from daytime images only, due to the constraint on the available depth dataset. More results can be found from our website\footnote{\url{http://www.isical.ac.in/~sanchayan_r/dehaze_icapr17}}.
\begin{figure}
    \centering     
    \begin{subfigure}[t]{0.49\linewidth}
        \includegraphics[width=\linewidth]{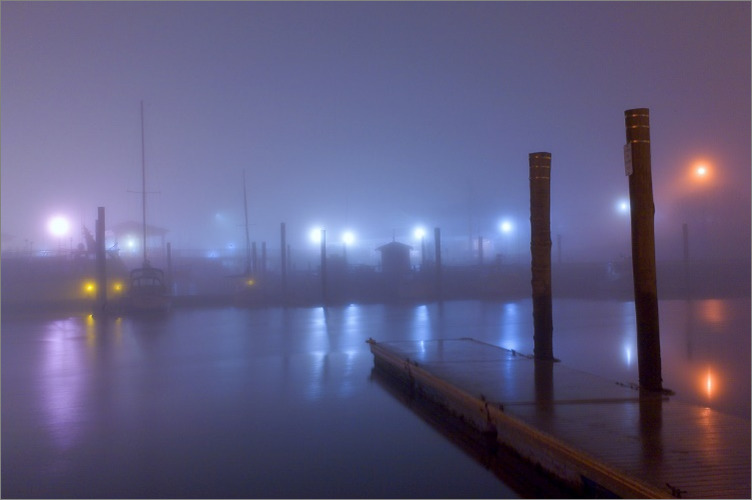}
        \caption{Input}
        \label{fig:inp}
    \end{subfigure}
    \begin{subfigure}[t]{0.49\linewidth}
        \includegraphics[width=\linewidth]{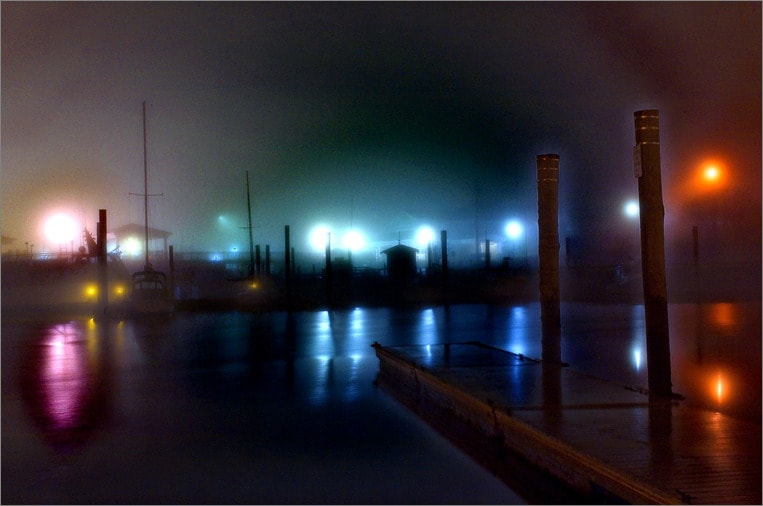}
        \caption{Our Method}
        \label{fig:cl}
    \end{subfigure}
    \caption{Failure case in night time image}
    \label{fig:res_night}
\end{figure}

\section{Conclusion}
\label{sec:conclusion}
Single image haze removal is one of the basic and important tasks to develop a robust and versatile computer vision system for object tracking, surveillance etc. In this paper, we have presented a method for single image dehazing using deep learning. The method assumes that the value of the transmittance is constant over a small patch and a constant atmospheric light over the whole image. The method uses a CNN to simultaneously estimate the transmittance value ($t(\mathbf{x})$) and environmental illumination ($A$) from a single patch by capturing the relationship between them as per the haze model (Eq. \eqref{eq:haze_patch}). The model is trained on a dataset consisting of hazy images, that was generated from clean images of NYU depth dataset, which contain depth maps for all the images. The transmittance map returned by the model contains some irregularities which is smoothed and the resultant transmittance map, along with the $A$ value is used to generate a haze-free image. The proposed method performs well for real outdoor scenes and the results are either better or comparable with other dehazing methods. The method can be extended for the night time images as we are estimating `$A$' for each patch separately.


\bibliographystyle{plain}
\bibliography{bibtex_file}

\end{document}